# Spike Timing Dependent Competitive Learning in Recurrent Self Organizing Pulsed Neural Networks Case Study: Phoneme and Word Recognition

Tarek BEHI[1], Najet AROUS[2] and Noureddine ELLOUZE[3]

[1]Electrical Engineering Department, National Engineering School of Tunis
Signal, Image and Pattern Recognition Research Unit,
Université Tunis El Manar BP 37, le Belvédère, Tunis 1002, Tunisia
tarekbehi@gmail.com

[2]Electrical Engineering Department, National Engineering School of Tunis
Signal, Image and Pattern Recognition Research Unit,
Université Tunis El Manar BP 37, le Belvédère, Tunis 1002, Tunisia
Najet.Arous@enit.rnu.tn

[3]Electrical Engineering Department, National Engineering School of Tunis
Signal, Image and Pattern Recognition Research Unit,
Université Tunis El Manar BP 37, le Belvédère, Tunis 1002, Tunisia
N.Ellouze@enit.rnu.tn

**Abstract**
Synaptic plasticity seems to be a capital aspect of the dynamics of neural networks. It is about the physiological modifications of the synapse, which have like consequence a variation of the value of the synaptic weight. The information encoding is based on the precise timing of single spike events that is based on the relative timing of the pre- and post-synaptic spikes, local synapse competitions within a single neuron and global competition via lateral connections. In order to classify temporal sequences, we present in this paper how to use a local hebbian learning, spike-timing dependent plasticity for unsupervised competitive learning, preserving self-organizing maps of spiking neurons. In fact we present three variants of self-organizing maps (SOM) with spike-timing dependent Hebbian learning rule, the Leaky Integrators Neurons (LIN), the Spiking_SOM and the recurrent Spiking_SOM (RSSOM) models. The case study of the proposed SOM variants is phoneme classification and word recognition in continuous speech and speaker independent.
*Keywords: Kohonen map, Leaky Integrators Neurons, Spiking_SOM, Recurrent Spiking_SOM, STDP, speech recognition.*

## 1. Introduction

Speech recognition, planning include the time parameter in an intrinsic manner. An obvious relation between the speech and time on the level of the phoneme is the scheduling of a sequence. It is obvious that a sequence of phonemes, carrying a message, cannot be modified by chance without enormously degrading the message. The modification of the pronunciation of a phoneme in context is however a fact known in voice recognition. This contextualization of the phonemes shows the need to know the sequence of the former phonemes, and sometimes posterior thus the problem of the context is clearly a problem of sequence and thus of time.

Current connectionist model are often static systems. They are powerful for patterns with no evolution in time, like in character recognition, but present some weaknesses if patterns involve a temporal component like in phoneme recognition. In our model the temporal information is taken into account by using spiking neurons, and by using time dependent mechanisms for synaptic plasticity.

Spike Timing Dependent Plasticity (STDP) is a temporally asymmetric form of Hebbian learning induced by tight temporal correlations between the spikes of pre- and postsynaptic neurons [1]. Times of emission of action potentials have an influence on synaptic plasticity. The STDP is based over the moments of emission of action potentials of the neurons pre- and postsynaptic. This form of plasticity was highlighted by the discovery of the existence of an action potentials retro propagate in dendrites [2], [3], and its influence on the weight of the synapse [4], [5]. The difference between time of emission of the presynaptic neuron $T_{pre}$ and time of postsynaptic emission of the neuron $T_{post}$ allowed the construction of temporal windows, defined the modification of the synaptic weight according to the difference $T = T_{pre} - T_{post}$. For the synapses, the order of the emissions is significant, while for the inhibiting synapses, only the temporal proximity is taken into account.

In the following, we present the self-organizing map of Kohonen (SOM) and the Spiking self-organizing map model (SSOM), thereafter, we explain the recurrent Spiking_SOM (RSSOM) and the Leaky Integrators Neurons model (LIN). Thereafter we expose Spike Timing Dependent Plasticity (STDP). Finally, we illustrate experimental results of the application of the three variants of SOM with Spike Timing Dependent Plasticity for phonemes and word recognition of TIMIT speech corpus

## 2. Spiking Self-Organizing Map

Self-organizing in networks is one of the most popular neural network fields [6], [7]. Such networks can learn to detect regularities and correlations in their input and adapt their future responses to that input accordingly [8], [9]. The neurons of competitive networks learn to recognize groups of similar input vectors. Self-organizing maps learn to recognize groups of similar input vectors in such a way that neurons physically near each other in the neuron layer respond to similar input vectors

A self-organizing map learns to categorize input vectors. It also learns the distribution of input vectors. Feature maps allocate more neurons to recognize parts of the input space where many input vectors occur and allocate fewer neurons to parts of the input space where few input vectors occur. Self-organizing maps also learn the topology of their input vectors.

An input vector $x \in \Re^n$ is compared with all $m_i$, in any metric; in practical applications, the smallest of the Euclidian distances is usually used to define the best-matching unit (BMU). The BMU is the neuron whose weight vector $m_i$ is closest to the input vector x determined by:

$$\|x - m_c\| = \min_i \{\|x - m_i\|\}, \forall\ i \in [1 .. n] \quad (1)$$

Where n is the number of map units and $\|x - m_i\|$ is a distance measure between x and $m_i$.

In the context of spiking neuron networks [10] we given a set S of m-dimensional input vectors $S = (s_1, ..., s_m)$ and a spiking neuron network with m input neurons and n output neurons, where each output neuron $v_j$ receives synaptic feedforward input from each input neuron $u_i$ with weight $w_{ij}$ and lateral synaptic input from each output neuron $v_k$, with weight $w_{kj}$. At every epoch of the learning procedure one sample is randomly chosen and the input neurons are made fire such that they temporally encode input vectors [10], [11].

In the algorithm proposed here, the winner is selected from the subpopulation of units that fire the quickest in one simulation step. After choosing a winner, learning is applied as follows. The afferent weights of a competitive neuron j are adapted in such a way as to maximize their similarity with the current input pattern i. A measure of the similarity is the difference between the postsynaptic potential $s_i$ that encodes the input stimulus and the connection weight $w_{ij}$. Furthermore, a spatial and a temporal neighborhood of the winner are created, such that only the neurons inside the S area and which have fired up until a reference time T (time interval of firing) are subject to learning.

The neighborhood-function, which is an important part of the SOM, is implemented here by the lateral connections among the output neurons. We assume that initially neurons which are topologically close together have strong excitatory lateral connections whereas remote neurons have strong inhibitory connections. This means that the firing of the winner neuron, say $v_k$, at time $t_k$ drives the firing times of neurons in the neighborhood of $v_k$ towards $t_k$, thus increasing the values they encode.

## 3. Recurrent spiking self organizing map

The recurrent Self-Organizing Map (RSOM) [12], [13] as an extension to the Self-Organizing Map (SOM), which allows storing certain information from the past input vectors. The information is stored in the form of difference vectors in the map units. The mapping that is formed during training has the topology preservation characteristic of the SOM. Recurrent SOM differs from the SOM only in its outputs. The outputs of the normal SOM are reseated to zero after presenting each input pattern and selecting best matching unit with the typical winner takes all strategy. Hence the map is sensitive only to the last input pattern. In the RSOM the sharp outputs are replaced with leaky integrator outputs, which once activated gradually lose their activity. The modeling of the outputs in RSOM is close to the behavior of natural neurons, which retain an electrical potential on their membranes with decay.
Thus, one can consider that the representation of time with this model is carried out by reintroducing in entry of the network the preceding state of the network. One notice in this case that time is used as a state which allows the successive reintroduction of the sequences the previous moment. This implicit representation of time takes into account that the aspect of order.

The use of impulsionnel neuron makes it possible to improve the taking into account of the temporal parameters like the duration and continuity.
The model presented in this part has the same principle that spiking self organizing map except that the choice of the BMU is defined by a difference vector in each unit of the map. The difference vector is included in the recurring bond. Thus, the memory stores a linear sum of the preceding vectors.
The difference vector $I_{(n)}$ in each unit of the map is defined there according to this equation:

$$y_i(t) = (1-\alpha) y_i(t-1) + \alpha(x(t) - m_i(t)) \quad (2)$$

Where $y_i(n)$ is the leaked difference vector in unit i, $0 < \alpha \leq 1$ is the leaking coefficient. $x(t)$ is the input vector and $m_i(t)$ is the weight vector of the unit i.

## 4. Leaky Integrators Neurons model

In this section we present an algorithm to train the temporal behavior of leaky integrator networks in the context of spiking neural networks. In a biological way, it is possible to convert temporal information into a static form by means of the low-pass filters, like those consisted the electric circuits with pair of resistance-condenser, as integrators [14]. This integrator is modeling with a simple mathematical model describing the change of potential inside a neuron by the use of temporal differential equations.

These dynamic neurons are called Leaky Integrators Neurons (LIN) [15].
In order to use leaky integrator units to create spiking neural network models for simulation experiments, a learning rule that works in continuous time is needed. The following formulation is motivated by [16] and describes how a backpropagation algorithm for leaky integrator units can be derived.

In this approach, the state of each neuron $i$ is represented by a membrane potential $P_i(T)$, which, is a function of the input $I(t)$ which measures the degree of matching between the neuron's weight vector and the current input vector.
The differential equation of a membrane potential is:

$$\frac{dP_i}{dt} = \eta P_i(t) + I_i(t) \quad (3)$$

Where $\eta < 0$. The input $I(t)$ of a neuron is a function of time.
Particularly, the discrete version of the equation (3), often written as:

$$P_i(t) = \lambda P_i(t-1) + I_i(t) \quad (4)$$

LIN stores the last activation of each neuron i by using the variable $a_i(t)$, this variable called the Leaky Integrators Neurons Potential.

$$a_i(t) = \lambda\, a_i(t-1) - \frac{1}{2} \| x(t) - w_i(t) \|^2 \quad (5)$$

where $0 \leq \lambda \leq 1$ is the memory depth constant, $x(t)$ is the input vector, and $w_i(t)$ is the weight vector of neuron *i*. Comparing equations (4) and (5), we find that $I_i(t) = -(\frac{1}{2}) \| x(t) - w_i(t) \|^2$.

## 5. Spike Timing Dependent Plasticity

We use a form of temporal synaptic plasticity corresponding to the STDP [17], [18]. The modelling of the STDP supposes the modification of the synaptic weight $\Delta W$ according to the difference $\Delta t = t_{pre} - t_{post}$, where $t_{pre}$ is the time of emission of pre-synaptic spike and $t_{post}$ the time of emission of post-synaptic spike.
There are several ways to modify the weights, of the synapse between neurons i and j, we can distinguish two variants from the STDP, the additive model and the multiplicative model.

With the additive model [19], [20], i.e. the variation of synaptic weight does not depend on the current weight, the optimal weights function of the difference between the pre-synaptic ($t_{pre}$) and post-synaptic ($t_{post}$) spike times is simply added to the synaptic weight.
The variation brought to the weight of the synapse is function of the difference between the pre-synaptic spike and the spike post-synaptic is updated by the rule:

$$\Delta W_{ij} = F(\Delta t) \quad (6)$$

Where $F(\Delta t)$ is the optimal weights function of the difference between the pre-synaptic ($t_{pre}$) and post-synaptic ($t_{post}$) spike times.

With the multiplicative model [19], [21], the effective modification of the weights relates to the way of limiting the weights so that those do not reach biologically unrealistic values. It depends on the difference between the pre-synaptic ($t_{pre}$) and post-synaptic ($t_{post}$) spike times and the current weight.
Panchev in [22], they suggest using the following multiplicative model:

$$\Delta W_{ij} = \begin{cases} w_{ij} + \eta F(\Delta t)(1 - w_{ij}) & \text{if } \Delta t > 0 \\ w_{ij} + \eta F(\Delta t) w_{ij} & \text{if } \Delta t < 0 \end{cases} \quad (7)$$

Where F ($\Delta$t) is the optimal weights function of the difference between the pre-synaptic ($t_{pre}$) and post-synaptic ($t_{post}$) spike times and η is the learning rate.

Soula in [23], they chose to modify the synaptic weights in the following way:

$$\Delta W_{ij} = \left\{ F(\Delta t) w_{ij} (1 - \frac{w_{ij}}{w_m}) \right\} \quad (8)$$

Where F ($\Delta$t) is the optimal weights function of the difference between the pre-synaptic ($t_{pre}$) and post-synaptic ($t_{post}$) spike times, $W_m$ represent the limit which the weights cannot exceed and η is the learning rate.

In our models we chose to modify the synaptic weight by using the following multiplicative model:

$$\Delta W_{ij} = \begin{cases} w_{ij} + \eta F(\Delta t)(x_i - w_{ij}) & \text{if } \Delta t > 0 \\ w_{ij} + \eta F(\Delta t) w_{ij} & \text{if } \Delta t < 0 \end{cases} \quad (9)$$

Where F ($\Delta$t) is the optimal weights function of the difference between the pre-synaptic ($t_{pre}$) and post-synaptic ($t_{post}$) spike times and η is the learning rate.

The optimal weights function is modeled by the equation:

$$F(\Delta t) = \begin{cases} A_+ e^{\Delta t / \tau_+} & \text{if } \Delta t < 0 \\ -A_- e^{\Delta t / \tau_-} & \text{if } \Delta t > 0 \end{cases} \quad (10)$$

Where $A_+$, $A_-$, $\tau_+$ and $\tau_-$ are the parameters of the law

## 6. Experimental results

### 6.1 The TIMIT speech database

The TIMIT acoustic-phonetic speech database was developed as a joint effort between researchers at the Massachusetts Institute of Technology (MIT), the Speech Research Institute (SRI) and Texas Instruments (TI) to evaluate factors related to acoustic variability in speech.

The database consists of three types of sentences that represent phonetic, contextual and speaker variations that are present in American English. The first type of sentences is the calibration sentences spoken by every speaker. These sentences were used to represent phonemes that would be spoken with the greatest amount of dialect variation (identified 'sa1 ' and ' sa2 '). The second type of sentences is the phonetically balanced and compact sentences spoken by several speakers. These sentences were created to represent the phonetic pairs in the English language (identified ' sx3 ' to ' sx452 '). Finally, the third type of sentences is randomly selected sentences to represent alternative occurrences of phonemes and to maximize the acoustic contexts, each sentence is marked only one time (identified ' si453 ' to ' si2343 ') [24], [25].

The database contains 6300 sentences, 10 sentences uttered by each of 630 speakers from 8 major dialect regions of the United States. The data were recorded at a sample rate of 16 KHz and a resolution of 16 bits.

### 6.2 Representation of speech data

We have used the TIMIT corpus for the purpose of developing and evaluating the proposed SOM variants for phonemes and word recognition in continuous speech and speaker independent.

The extraction of the best parametric representation of acoustic signals is an important task to produce a better recognition performance. The efficiency of this phase is important for the next phase since it affects its behavior.

A wide range of possibilities exist for parametrically representing the speech signal for the speaker recognition task, such as Linear Prediction Coding (LPC), Mel-Frequency Cepstrum Coefficients (MFCC), and others. MFCC is perhaps the best known and most popular.

Mel-Frequency Cepstral Coefficients (MFCC) is the standard preprocessing technique in the field of speech recognition. The MFCC are calculated as follows: First, the sample data is windowed using a hamming window and a FFT is computed, thereafter, the magnitude is run through the mel-scale (is a non-linear transformation of the frequency domain to model the human selectivity to certain frequency bands.) filter bank and the $\log_{10}$ of these values is computed, finally, a cosine transform is applied to reduce the correlation among the individual features. The result is the cepstrum.

Speech utterance was sampled at a sampling rate of 16 KHz using 16 bits quantization. Speech frames are filtered by a first order filter whose transfer function is:

$$H(z) = 1 - a.z^{-1}, \quad 0.9 \leq a \leq 1.0 \quad (11)$$

Where $z^{-1}$ is the delay operator. In our experiments, a is chosen to be 0.95.
After the pre-emphasis, speech data consists of a large amount of samples that present the original utterance.

Windowing is introduced to effectively process these samples. This is done by regrouping speech data into several frames. In our system, a 256 sample window that could capture 16 ms of speech information is used. To prevent information lost during the process, an overlapping factor of 50% is introduced between adjacent frames.

After regrouping, each individual frame needs to be further pre-processed to minimize signal discontinuities at the beginning and at the end of each frame. A commonly used technique is to multiply the signal data with the hamming function. The earlier has smoothing effects at edges of the filter. This function can be described by the following equation:

$$h(n) = 0.54 - 0.46 * \cos(\frac{2\pi n}{N-1}) \quad 0 \leq n \leq N, \ N > 1 \quad (12)$$

Where n is the sample number and N is the total number of samples per window. In our case, N is 256.

Thereafter, mel frequency cepstral analysis was applied to extract the 12 mel cepstrum coefficients. The mel scale is an equi-pitch scale describing the subjective and perceptual response to frequency of human listener. The implemented neural networks are trained by presenting them with 12 input values from 9 frames selected at the middle of each phoneme.

### 6.3 Results and discussions

We have implemented the Kohonen model based on sequential learning and the proposed SOM variants. The realized system is composed of three main components [26], [27]: a pre-processor sounds and producing mel cepstrum vectors. The sound input space is composed by 12 mel cepstrum coefficients each 16 ms frame. 9 frames are selected at the middle of each phoneme to generate data vectors. The second component is a competitive learning module. The third component is a phoneme recognition module.

All maps are trained for 80 iterations using a train data set. For all maps, the learning rate decrease linearly from 0.9 to 0.05. The radius width decrease also linearly from half the diameter of the lattice to one. All maps of the same size have same initial conditions (that is the same $m_i(0)$). The neural lattice was bidimensional.

For phoneme recognition, we have used the New England dialect region (DR1) composed of 31 male and 18 female. The corpus contains 18547 phonetic units. Each phonetic unit is represented by 9 frames selected at the middle of each phoneme to generate data vectors. Training has been made on phonemes for the seven macro classes of TIMIT database. Table 1 shows the list of phonemes of each macro-class of TIMIT data base.

Table 1: List of phonemes of phonemic classes of each macro-class

| Macro-class | Phonemes |
|---|---|
| affricates | /jh/, /ch/ |
| stops | /b/, /d/, /g/, /p/, /t/, /k/, /dx/, /q/, /bcl/, /dcl/, /gcl/, /pcl/, /tcl/, /kcl/ |
| Others | /pau/, /epi/, /h#/ |
| Nasals | /m/, /n/, /ng/, /em/, /en/, /eng/, /nx/ |
| Semi-vowels | /l/, /r/, /w/, /y/, /hh/, /hv/, /el/ |
| Fricatives | /s/, /sh/, /z/, /zh/, /f/, /th/, /v/, /dh/ |
| vowels | /iy/, /ih/, /eh/, /ey/, /ae/, /aa/, /aw/, /ay/, /ah/, /ao/, /oy/, /ow/, /uh/, /uw/, /ux/, /er/, /ax/, /ix/, /axr/, /axh/ |

From table 2 RSSOM and LIN obtained an improvement of the classification rate of 10 % in comparison with SOM. RSSOM and LIN reach good classification rates in order to 96% for /affricate/.

The results in Table 2 show that Leaky Integrators Neurons provide best classification rate for /others/ and /Semi-vowels/ in order to 93%

Table 2: General recognition rates of the 7 macro-classes of TIMIT data base (training set)

| Macro-class | SOM | SSOM | RSSOM | LIN |
|---|---|---|---|---|
| Affricates | 79.59 | 85.71 | 95.91 | 95.91 |
| Fricatives | 67.67 | 71.47 | 74.07 | 77.33 |
| Others | 87.99 | 89.96 | 91.77 | 93.09 |
| Nasals | 71.83 | 84.71 | 84.71 | 84.71 |
| Stops | 47.05 | 53.29 | 54.20 | 56.38 |
| Semi-vowels | 84.61 | 91.22 | 91.98 | 93.06 |
| Vowels | 51.81 | 60.42 | 62.15 | 66.30 |
| Average | 70.07 | 78.11 | 79.25 | 80.96 |

According to table 3, LIN obtained an improvement of the classification rate of 12% in comparison with SOM. LIN reaches good classification rates in order to 93% for /Nasals/.

For word recognition, we have used the New England dialect region (DR1) composed of 24 male and 14 female. The corpus contains 7380 word units for training. Each word unit is represented by 9 frames selected at the middle of each word to generate data vectors. Training has been made on words for ten sentences of TIMIT database. The sentences and words can be found in table 4.

Table 3: General recognition rates of the 7 macro-classes of TIMIT data base (test set)

| Macro-class | SOM | SSOM | RSSOM | LIN |
|---|---|---|---|---|
| Affricates | 74.64 | 83.25 | 84.68 | 85.16 |
| Fricatives | 61.61 | 71.47 | 73.26 | 77.26 |
| Others | 64.98 | 72.33 | 75.47 | 75.39 |
| Nasals | 79.51 | 87.64 | 90.40 | 92.84 |
| Stops | 43.04 | 47.19 | 49.94 | 52.23 |
| Semi-vowels | 78.42 | 83.76 | 87.84 | 89.59 |
| Vowels | 51.09 | 58.79 | 62.93 | 64.06 |
| Average | 64.75 | 72.06 | 74.93 | 76.64 |

Table 4: List of words of each sentence

| Sentences | Words |
|---|---|
| SA1 | {'she'} {'had'} {'your'} {'dark'} {'suit'} {'in'} {'greasy'} {'wash'} {'water'} {'all'} {'year'} |
| SA2 | {'don't'} {'ask'} {'me'} {'to'} {'carry'} {'an'} {'oily'} {'rag'} {'like'} {'that'} |
| SI1027 | {'even'} {'then'} {'if'} {'she'} {'took'} {'one'} {'step'} {'forward'} {'he'} {'could'} {'catch'} {'her'} |
| SX56 | {'academic'} {'aptitude'} {'guarantees'} {'your'} {'diploma'} |
| SI1608 | {'She'} {'smiled'} {'and'} {'teeth'} {'gleamed'} {'in'} {'her'} {'beautifully'} {'modeled'} {'olive'} {'face'} |
| SI1377 | {'as'} {'these'} {'maladies'} {'overlap'} {'so'} {'must'} {'the'} {'cure'} |
| SI1244 | {'the'} {'sculptor'} {'looked'} {'at'} {'him'} {'bugeyed'} {'and'} {'amazed'} {'angry'} |
| SX395 | {'i'} {'took'} {'her'} {'word'} {'for'} {'it'} {'but'} {'is'} {'she'} {'really'} {'going'} {'with'} {'you'} |
| SI1621 | {'Now'} {'he'} {'ll'} {'choke'} {'for'} {'sure'} |
| SX111 | {'His'} {'sudden'} {'departure'} {'shocked'} {'the'} {'cast'} |

According to table 5, LIN provides the best classification rate in order to 70%. With LIN we obtained an improvement of the classification rate in order to 12 % in comparison with SOM.
It is also noticed that for the sentence 'SA1', the three variants (SSOM, RSSOM and LIN) have capacities of roughly similar recognition.

Table 5: Sentence SA1 recognition rates

| Word | SOM | SSOM | RSSOM | LIN |
|---|---|---|---|---|
| she | 80.11 | 77.48 | 80.70 | 84.50 |
| had | 73.97 | 82.45 | 81.28 | 80.99 |
| your | 36.47 | 42.85 | 50.75 | 44.68 |
| dark | 65.78 | 80.11 | 78.07 | 90.93 |
| suit | 60.78 | 69.00 | 69.59 | 70.17 |
| in | 74.08 | 78.96 | 78.04 | 73.87 |
| greasy | 31.28 | 56.14 | 49.41 | 51.75 |
| wash | 46.78 | 44.73 | 50.58 | 65.20 |
| water | 37.71 | 45.90 | 53.21 | 52.04 |
| all | 74.26 | 76.02 | 74.26 | 77.48 |
| year | 55.26 | 70.76 | 71.34 | 76.60 |
| Average | 57.85 | 65.89 | 67.04 | 69.90 |

Table 6 shows that SSOM provide the best recognition accuracy in order to 63.40%. LIN provides best rate for the word /rag/ in order to 75.55%.
With SOM model we can't recognize some words like /like/, recognition rates in the range of 20%.

Table 6: Sentence SA2 recognition rates

| Word | SOM | SSOM | RSSOM | LIN |
|---|---|---|---|---|
| don't | 67.30 | 68.57 | 68.88 | 72.38 |
| ask | 46.98 | 59.68 | 58.73 | 62.53 |
| me | 61.05 | 67.71 | 71.57 | 60.00 |
| to | 44.63 | 57.04 | 49.66 | 66.44 |
| carry | 57.14 | 73.65 | 66.34 | 53.65 |
| an | 47.95 | 60.54 | 59.52 | 55.78 |
| oily | 58.41 | 67.93 | 69.52 | 66.66 |
| rag | 62.22 | 72.06 | 66.34 | 65.07 |
| like | 19.36 | 37.14 | 43.49 | 28.57 |
| that | 72.06 | 73.96 | 72.06 | 75.55 |
| Average | 52.30 | 63.40 | 61.12 | 59.79 |

According to table 7, LIN model provide best classification rate in order to 97.24% in training set. With LIN we obtained an improvement of the classification rate in order to 30 % in comparison with SOM.
With SOM model we can't recognize some words like /for/ and /is/, recognition rates in the range of 0 and 20%, on the other hand the Leaky Integrators Neurons (LIN) provide a higher value of recognition rate (in the range of 90 and 100%).

Table 7: Sentence SX395 recognition rates

| Word | SOM | SSOM | RSSOM | LIN |
|---|---|---|---|---|
| i | 100 | 100 | 100 | 100 |
| took | 88.88 | 100 | 66.66 | 100 |
| her | 77.77 | 100 | 100 | 100 |
| word | 77.77 | 100 | 100 | 100 |
| for | 0.00 | 55.55 | 100 | 88.88 |
| it | 60.00 | 80.00 | 80.00 | 100 |
| but | 77.77 | 88.88 | 88.88 | 100 |
| is | 20.00 | 100 | 100 | 80.00 |
| she | 100 | 100 | 100 | 100 |
| really | 44.44 | 77.77 | 88.88 | 100 |
| going | 55.55 | 100 | 100 | 100 |
| with | 55.55 | 88.88 | 88.88 | 88.88 |
| you | 77.77 | 88.88 | 100 | 100 |
| Average | 66.05 | 90.82 | 93.57 | 97.24 |

According to table 8, Leaky Integrators Neurons (LIN) provides the best classification rate in order to 96.25%. With LIN we obtained an improvement of the classification rate in order to 19 % in comparison with SOM in training set.

A higher value of recognition rate for the words /looked/ and / bugeyed/ means a better performance of variant SOM.

Ttable 8: Sentence SI1244 recognition rates

| Word | SOM | SSOM | RSSOM | LIN |
|---|---|---|---|---|
| the | 87.50 | 87.50 | 87.50 | 100 |
| sculptor | 100 | 100 | 100 | 100 |
| looked | 66.66 | 88.88 | 88.88 | 100 |
| at | 88.88 | 100 | 88.88 | 100 |
| him | 77.77 | 100 | 55.55 | 100 |
| bugeyed | 44.44 | 100 | 100 | 100 |
| and | 44.44 | 66.66 | 100 | 66.66 |
| amazed | 88.88 | 88.88 | 88.88 | 100 |
| angry | 100 | 100 | 100 | 100 |
| Average | 77.50 | 92.50 | 90.00 | 96.25 |

From table 9 with the proposed SOM variants we obtained an improvement of the classification rate of 13 % in comparison with SOM.

The SOM variants reach good classification rates in order to 91% in training set.

Table 9: Sentence SX56 recognition rates

| Word | SOM | SSOM | RSSOM | LIN |
|---|---|---|---|---|
| academic | 66.66 | 100 | 100 | 88.88 |
| aptitude | 100 | 100 | 88.88 | 100 |
| guarantees | 33.33 | 55.55 | 66.66 | 66.66 |
| your | 100 | 100 | 100 | 100 |
| diploma | 100 | 100 | 100 | 100 |
| Average | 78.57 | 90.47 | 90.47 | 90.47 |

From table 10 RSSOM, LIN and Spiking_SOM provide best classification accuracy in comparison with SOM. The variant LIN reaches good classification rates in order to 97.22%. However, this model reaches good recognition rates (in the range of 90 and 100%).
With LIN we obtained an improvement of the classification rate in order to 15 % in comparison with SOM.

For the word /one/, SOM has the low recognition rate in order to 10%, but with the proposed variants models we obtained a recognition rate in order to 100%, this result prove the stability and performance of the SOM variants.

Table 10: Sentence SI1027 recognition rates

| Word | SOM | SSOM | RSSOM | LIN |
|---|---|---|---|---|
| even | 88.88 | 100 | 77.77 | 100 |
| then | 100 | 100 | 100 | 100 |
| if | 77.77 | 88.88 | 100 | 100 |
| she | 100 | 66.66 | 88.88 | 77.77 |
| took | 100 | 100 | 77.77 | 100 |
| one | 11.11 | 100 | 100 | 100 |
| step | 88.88 | 100 | 100 | 100 |
| forward | 100 | 100 | 100 | 100 |
| he | 66.66 | 100 | 100 | 100 |
| could | 88.88 | 100 | 88.88 | 88.88 |
| catch | 77.77 | 100 | 100 | 100 |
| her | 88.88 | 100 | 100 | 100 |
| Average | 82.40 | 96.29 | 94.44 | 97.22 |

According to table 11, Leaky Integrators Neurons (LIN) provides the best classification rate in order to 96.25%. With LIN we obtained an improvement of the classification rate in order to 19 % in comparison with SOM in training set.
A higher value of recognition rate for the words /looked/ and / bugeyed/ means a better performance of variant SOM.

Table 11: Sentence SI1244 recognition rates

| Word | SOM | SSOM | RSSOM | LIN |
|---|---|---|---|---|
| the | 87.50 | 87.50 | 87.50 | 100 |
| sculptor | 100 | 100 | 100 | 100 |
| looked | 66.66 | 88.88 | 88.88 | 100 |
| at | 88.88 | 100 | 88.88 | 100 |
| him | 77.77 | 100 | 55.55 | 100 |
| bugeyed | 44.44 | 100 | 100 | 100 |
| and | 44.44 | 66.66 | 100 | 66.66 |
| amazed | 88.88 | 88.88 | 88.88 | 100 |
| angry | 100 | 100 | 100 | 100 |
| Average | 77.50 | 92.50 | 90.00 | 96.25 |

According to table 12, Leaky integrators neurons (LIN) and recurrent spiking SOM (RSSOM) provide best classification rate in order to 100% in training set.
SOM model gives the weak result for the word /so/ in order to 44%, on the other hand, all the other models provide a higher value of recognition rate in order to 100%.

Table 12: Sentence SI1377 recognition rates

| Word | SOM | SSOM | RSSOM | LIN |
|---|---|---|---|---|
| as | 100 | 100 | 100 | 100 |
| these | 100 | 100 | 100 | 100 |
| maladies | 66.66 | 100 | 100 | 100 |
| overlap | 100 | 100 | 100 | 100 |
| so | 44.44 | 88.88 | 100 | 100 |
| must | 100 | 77.77 | 100 | 100 |
| the | 100 | 100 | 100 | 100 |
| cure | 88.88 | 100 | 100 | 100 |
| Average | 87.50 | 95.83 | 100 | 100 |

According to table 13, LIN provide best classification rate in order to 100% in training set. With LIN we obtained an improvement of the classification rate in order to 14 % in comparison with SOM.

It is also noticed that for the sentence 'SI1621', the variants SSOM and RSSOM have capacities of roughly similar recognition.

According to table 14, the variant RSSOM provide best classification rate in order to 96% in training set.

With SOM model we can't recognize some words like /gleamed/, recognition rates (in the range of 20). On the other hand the RSSOM provide a higher value of recognition rate (in the range of 90 and 100%).

Table 13: Sentence SI1621 recognition rates

| Word | SOM | SSOM | RSSOM | LIN |
|---|---|---|---|---|
| now | 100 | 100 | 100 | 100 |
| he'll | 77.77 | 100 | 100 | 100 |
| choke | 88.88 | 100 | 88.88 | 100 |
| for | 100 | 88.88 | 100 | 100 |
| sure | 66.66 | 100 | 100 | 100 |
| Average | 86.66 | 97.77 | 97.77 | 100 |

Table 14: Sentence SI1608 recognition rates

| Word | SOM | SSOM | RSSOM | LIN |
|---|---|---|---|---|
| she | 100 | 100 | 88.88 | 100 |
| smiled | 100 | 100 | 100 | 100 |
| and | 100 | 100 | 100 | 100 |
| teeth | 66.66 | 77.77 | 88.88 | 77.77 |
| gleamed | 22.22 | 88.88 | 88.88 | 66.66 |
| in | 77.77 | 88.88 | 88.88 | 100 |
| her | 88.88 | 100 | 100 | 100 |
| beautifully | 88.88 | 100 | 100 | 100 |
| modeled | 55.55 | 100 | 100 | 100 |
| olive | 100 | 100 | 100 | 100 |
| face | 100 | 77.77 | 100 | 100 |
| Average | 82.82 | 93.93 | 95.95 | 94.94 |

According to table 15, SSOM and RSSOM models provide best classification rate in order to 100% in training set.

With RSSOM and SSOM we obtained an improvement of the classification rate in order to 10 % in comparison with SOM.

Table 15: Sentence SX111 recognition rates

| Mots | SOM | SSOM | RSSOM | LIN |
|---|---|---|---|---|
| his | 77.77 | 100 | 100 | 100 |
| sudden | 100 | 100 | 100 | 100 |
| departure | 100 | 100 | 100 | 100 |
| shocked | 100 | 100 | 100 | 100 |
| the | 66.66 | 100 | 100 | 88.88 |
| cast | 100 | 100 | 100 | 100 |
| Average | 90.74 | 100 | 100 | 98.14 |

## 7. Conclusions

In this paper, we have proposed a new approach of Kohonen's algorithm for spiking neurons using temporal coding applied to speech recognition. The information encoding is based on the precise timing of single spike events. The idea is to provide a competitive learning algorithm, spike-timing dependent synaptic plasticity based on the relative timing of the pre- and post-synaptic spikes, local synapse competitions within a single neuron and global competition via lateral connections.

The case study of the proposed spiking self organizing map models is phoneme and word recognition in continuous speech and speaker independent.
The proposed SOM variants provide best classification rates of phonemes and word classification in comparison with the basic SOM model.

The LIN provides the best general recognition rates of the 7 macro-classes of TIMIT data base in order to 80.96% in training set and 76.64% in test set.
The LIN provides the best general recognition rates of sentences of TIMIT data base in order to 95% in training set.

As a future work, we propose to implement a cooperative system of self-organizing spiking neurons (CoopSOSN) for phoneme and word recognition. The CoopSOSN is based on the association of different SOM variants of unsupervised and Spike-Timing Dependent Hebbian learning algorithms. The objective of such system is to create a cooperative system based on different competitive learning algorithms in order to improve classification rates.

We suggest also to hybridize SOM and genetic algorithm on one hand to fine tune SOM parameters and on the other hand for training data set input in the objective to ameliorate recognition rates

**Tarek BEHI** received the MS degree in electrical engineering (signal processing) from National School of Engineer of Tunis (ENIT–Tunisia), is currently working towards the Ph.D. degree in electrical engineering (signal processing) from ENIT. He is currently a computer science assistant in the informatics department at FST, Tunisia. Her research interests include speech recognition, spiking neural networks, hierarchical neural networks and evolutionary neural networks.

**Najet AROUS** received computer science engineering degree from Ecole Nationale des Sciences d'Informatique, Tunis, Tunisia, the MS degree in electrical engineering (signal processing) from the National School of Engineer of Tunis (ENIT–Tunisia), Tunisia, the Ph.D. degree in electrical engineering (signal processing) from ENIT. She is currently a computer science assisting master in the computer science department at FSM, Tunisia. Her research interests include scheduling optimization, speech recognition and evolutionary neural networks.

**Noureddine ELLOUZE** received a Ph.D. degree in 1977 from l'Institut National Polytechnique at Paul Sabatier University (Toulouse-France), and Electronic Engineer Diploma from ENSEEIHT in 1968 at the same University.
In 1978, Dr. Ellouze joined the Department of Electrical Engineering at the National School of Engineer of Tunis (ENIT–Tunisia), as ASSISTANT PROFESSOR in statistic, electronic, signal processing and computer architecture. In 1990, he became Professor in signal processing; digital signal processing and stochastic process. He has also served as director of electrical department at ENIT from 1978 to 1983. General manager and President of the Research Institute on Informatics and Telecommunication IRSIT from 1987-1990, and President of the Institut in 1990-1994. He is now Director of Signal Processing Research Laboratory LSTS at ENIT, and is in charge of Control and Signal Processing Master degree at ENIT.
Pr Ellouze is IEEE fellow since 1987; he directed multiple Masters and Thesis and published over 200 scientific papers both in journals and proceedings. He is chief editor of the scientific journal Annales Maghrébines de l'Ingénieur. His research interest include neural networks and fuzzy classification, pattern recognition, signal processing and image processing applied in biomedical, multimedia, and man machine communication.